\documentclass[pdflatex,sn-mathphys-num]{sn-jnl}% Math and Physical Sciences Numbered Reference Style
\usepackage{graphicx}%
\usepackage{multirow}%
\usepackage{amsmath,amssymb,amsfonts}%
\usepackage{amsthm}%
\usepackage{mathrsfs}%
\usepackage[title]{appendix}%
\usepackage{xcolor}%
\usepackage{textcomp}%
\usepackage{manyfoot}%
\usepackage{booktabs}%
\usepackage{algorithm}%
\usepackage{algorithmicx}%
\usepackage{algpseudocode}%
\usepackage{listings}%
\usepackage{array}%
\usepackage{pifont}%
\theoremstyle{thmstyleone}%
\newcolumntype{C}[1]{>{\centering\arraybackslash}p{#1}}
\theoremstyle{thmstyletwo}%

\theoremstyle{thmstylethree}%

\begin{document}

\title[Article Title]{VTK-Bench: A Diagnostic Benchmark for Distinguishing Language Priors and Text Modal Bias in Visual Language Models}


\author[1,2]{\fnm{Yao} \sur{Lu}}
% \email{yaolu.zjut@gmail.com}

\author[1,2]{\fnm{Yuhang} \sur{Zheng}}

\author[1,2]{\fnm{Yaxin} \sur{Gao}}

\author[2]{\fnm{Zhe} \sur{Tang}}

\author[1,2]{\fnm{Zhaowei} \sur{Zhu}}

\author[3]{\fnm{Jiaheng} \sur{Wei}}

\author[1,2]{\fnm{Yunhao} \sur{Wang}}

\author*[1,2]{\fnm{Shanqing} \sur{Yu}}\email{shanqingyu1984@gmail.com}

\author[1,2]{\fnm{Qi} \sur{Xuan}}

\affil*[1]{\orgdiv{Institute of Cyberspace Security}, \orgname{Zhejiang University of Technology}, \orgaddress{\country{China}}}

\affil[2]{\orgdiv{Binjiang Institute of Artificial Intelligence}, \orgname{Zhejiang University of Technology}, \orgaddress{\country{China}}}

\affil[3]{\orgname{Hong Kong University of Science and Technology (Guangzhou)}, \orgaddress{\country{China}}}


\abstract{Visual language models (VLMs) have demonstrated exceptional capabilities across various multimodal tasks, but they remain susceptible to hallucinations, undermining their reliability. Existing hallucination benchmarks primarily focus on identifying hallucinations and describing their manifestations, often neglecting their underlying causes. Specifically, multimodal hallucinations can stem from two distinct failure modes: text modality bias (TMB), reflecting an over-reliance on misleading text input; and language prior (LP), reflecting a reliance on previously learned knowledge. To differentiate these failure modes, we propose VTK-Bench, a benchmark that explicitly separates three information sources: visual information (V), textual information (T), and prior knowledge (K). By establishing mutually exclusive answers corresponding to V, T, and K, VTK-Bench enables fine-grained attribution analysis of model responses. Furthermore, we introduce discrete attribution and soft attribution metrics to characterize information source preferences. Extensive experiments on $11$ representative open-source VLMs show that LP is more prevalent than TMB in most evaluated models. Notably, TMB decreases substantially with model scaling, whereas LP exhibits relatively limited improvement and remains the dominant failure mode. These findings highlight the importance of different information source preferences for a deeper understanding of multimodal hallucinations.}




\maketitle

\section{Introduction}\label{sec1}

Recent advances in large language models~\cite{singh2025openai,lu2025structural,lu2026reassessing} (LLMs) and visual representation learning~\cite{he2020momentum} have significantly accelerated the development of VLMs. By integrating powerful visual encoders with large language models, modern VLMs have demonstrated remarkable capabilities across a wide range of tasks, including visual question answering~\cite{antol2015vqa}, image captioning~\cite{hossain2019comprehensive}, visual grounding~\cite{xu2024vlm} and complex visual reasoning~\cite{jiang2026vlm}. Despite their superior performance, VLMs remain susceptible to multimodal hallucination~\cite{huang2025survey}, particularly when visual evidence is inconsistent with textual descriptions or previously acquired knowledge. Therefore, understanding the causes of this inconsistency is crucial for improving the reliability and credibility of VLMs.

Existing hallucination evaluation methods~\cite{liu2024survey,chen2026survey} primarily focus on identifying hallucinations and characterizing their manifestations. While these methods provide effective tools for detecting inconsistencies between model output and visual evidence, they often overlook the root causes of hallucinations. In fact, seemingly same hallucinations can stem from drastically different causes. For example, consider a VLM presented with an image depicting a drone equipped with eight propellers, yet generating a response stating that the drone has only four. Traditional hallucination detection methods~\cite{manakul2023selfcheckgpt} would identify this response as a hallucination due to its inconsistency with the visual evidence. However, this alone doesn't reveal the root cause of the incorrect answer. At least three explanations exist. First, the model may be unable to accurately count the propellers, indicating insufficient visual perception. Second, the model may have recognized the visual content but was misled by incorrect descriptions in the accompanying textual question. Third, the model may possess the necessary perceptual ability but tends to rely on its previously learned perception that "a typical drone has four propellers," even when the image explicitly contradicts this expectation. While these situations may all lead to the same incorrect predictions, they stem from different failure mechanisms. Treating them as a single hallucination phenomenon obscures their underlying causes and limits our understanding of model reliability.

\begin{table*}[t]
    \centering
    \caption{\textbf{Comparison of representative VLM hallucination benchmarks.} TMB Eval. = textual modality bias evaluation; LP Eval. = language prior evaluation; TMB–LP Disent. = disentanglement of textual modality bias and language prior; Cap. Verif. = capability verification; Dual Attr. = dual attribution using discrete and soft metrics; Scale = dataset size. \ding{51} indicates full coverage of the corresponding criterion; $\Delta$ indicates partial coverage; and \ding{55} indicates no coverage. PhD-icc denotes the incorrect-context subset of PhD, which is included because its evaluation setting is closely related to our TMB evaluation.}
    \label{tab:benchmark_comparison}
    \small
    \setlength{\tabcolsep}{4pt}
    \renewcommand{\arraystretch}{1.15}
    \resizebox{\textwidth}{!}{
    \begin{tabular}{lccccccc}
        \toprule
        Benchmark
        & TMB Eval.
        & LP Eval.
        & TMB--LP Disent.
        & Cap. Verif.
        & Dual Attr.
        & Scale \\
        \midrule
        POPE~\cite{li2023evaluating}
        & \ding{55} & \ding{55} & \ding{55} & \ding{55} & \ding{55} & $3,000$ \\

        MMBench~\cite{liu2024mmbench}
        & \ding{55} & \ding{55} & \ding{55} & \ding{55} & \ding{55} & $2,974$ \\

        PhD-icc~\cite{liu2025phd}
        & \ding{51} & $\Delta$ & \ding{55} & $\Delta$ & \ding{55} & $33,688$ \\

        HallusionBench~\cite{guan2024hallusionbench}
        & \ding{55} & $\Delta$ & \ding{55} & $\Delta$ & \ding{55} & $1,129$ \\
        
        VLind-Bench~\cite{lee2025vlind}
        & \ding{55} & \ding{51} & \ding{55} & \ding{51} & \ding{55} & $302$ \\

        ViLP~\cite{luo2024probing}
        & \ding{55} & \ding{51} & \ding{55} & \ding{55} & \ding{55} & $300$ \\

        \midrule
        \textbf{VTK-Bench (Ours)}
        & \ding{51} & \ding{51} & \ding{51} & \ding{51} & \ding{51} & \textbf{300} \\
        \bottomrule
    \end{tabular}
    }
\end{table*}

To better characterize these failures, we distinguish two fundamentally different forms of hallucination: Text Modality Bias (TMB) and Language Prior (LP). TMB refers to the tendency of VLMs to adopt misleading textual information rather than contradictory visual evidence, reflecting their over-reliance on textual input. In contrast, LP refers to the tendency of VLMs to rely on previously learned knowledge rather than visual evidence presented in the current context. Existing hallucination benchmarks, such as POPE~\cite{li2023evaluating}, AMBER~\cite{wang2023amber} and HallusionBench~\cite{guan2024hallusionbench}, primarily focus on detecting hallucinations and evaluating visual consistency. Although recent benchmarks (Table~\ref{tab:benchmark_comparison}), such as VLind-Bench~\cite{lee2025vlind} and ViLP~\cite{luo2024probing}, have investigated the influence of language priors, they do not explicitly distinguish the effects of external textual interference and previously learned knowledge within a unified evaluation framework.

To address this limitation, we introduce VTK-Bench, a benchmark designed to distinguish TMB and LP by explicitly separating three information sources: visual information (V), textual information (T), and prior knowledge (K). Specifically, we construct a dataset comprising $300$ samples and develop a controlled evaluation protocol that combines consistency verification, prior knowledge verification and cross-modal conflict evaluation. By designing evaluation scenarios in which V, T, and K correspond to mutually exclusive answers, VTK-Bench is able to perform fine-grained attribution of different information sources to which the model responds. In addition, we introduce discrete and soft attribution indicators to characterize model preferences from two perspectives: final response selection and word element probability.

We conduct extensive evaluations on $11$ representative open-source VLMs, covering model sizes from $2$B to $32$B parameters and including both dense and mixture-of-experts architectures. Our experiments reveal several important findings. Firstly, LP is more prevalent than TMB across all models evaluated. Secondly, as the scale of the Qwen3-VL model increases, the visual realism is improved, TMB decreases significantly, and LP shows a non-monotonic change trend. Thirdly, discrete attribution and soft attribution show high consistency in model rankings, while the quantitative difference between them reveals the model's preference for different information sources. These findings highlight the importance of distinguishing TMB and LP for a more comprehensive understanding of multimodal hallucinations.

To summarize, our main contributions are three-fold:

\begin{itemize}
    \item \textbf{Redefining the diagnostic problem of multimodal hallucinations.} We distinguish two fundamentally different failure modes, Text Modality Bias and Language Prior, by identifying their underlying information sources.
     \item \textbf{A novel benchmark for multimodal hallucinations.} We construct a new dataset comprising $300$ samples, enabling fine-grained evaluation of Text Modality Bias and Language Prior in VLMs.
      \item \textbf{Extensive evaluations and insightful findings.} We conduct comprehensive experiments on $11$ representative VLMs, revealing different hallucination patterns across model scales and families. Notably, Text Modality Bias decreases substantially with model scaling, while Language Prior remains the dominant failure mode.
\end{itemize}

\section{Related Work}\label{sec2}

\subsection{Multimodal Hallucination}\label{subsec2.1}

Vision-Language Models (VLMs) remain susceptible to multimodal hallucinations, often generating content that is inconsistent with visual inputs. Existing research on this issue mainly focuses on the detection and assessment of hallucinations. For example, CHAIR~\cite{rohrbach2018object} evaluates the visual fidelity of the model by detecting object hallucinations appearing in image descriptions; POPE~\cite{li2023evaluating} converts object hallucination evaluation into a binary question answering task; AMBER~\cite{wang2023amber} builds a comprehensive evaluation benchmark covering multiple hallucination types; HallusionBench~\cite{guan2024hallusionbench} evaluates the limitations of the model in visual perception and logical reasoning by designing challenging visual question and answer samples.

The aforementioned research has advanced the development of multimodal hallucination assessment, but its focus is mainly on identifying whether the model produces hallucinations and characterizing the specific manifestations of hallucinations. Existing assessments lack further distinction regarding which information sources the model relies on in cross-modal information conflict scenarios. In particular, when the model output is inconsistent with visual evidence, the reason may be interference from external textual information or it may stem from the model's own linguistic prior bias. Without distinguishing between these two types of information sources, it is difficult to accurately analyze the role of different factors in multimodal illusions, and it is also difficult to further reveal the decision-making tendency of the model under cross-modal information conflict.

\subsection{Cross-modal Bias and Language Prior}\label{subsec2.2}

Cross-modal bias and language prior are important factors affecting the visual fidelity of multi-modal models. Existing research has pointed out that VLMs may show a bias towards specific modalities when processing graphic information, such as over-reliance on verbal information and ignoring visual evidence~\cite{peng2022balanced}. Related research further analyzed the imbalance between modalities from the perspective of model training and internal representation, and proposed to alleviate modal bias through gradient modulation and other methods~\cite{wang2020makes}.

Meanwhile, language prior are also a common influencing factor in visual question answering tasks. VQA-CP~\cite{agrawal2018don} assesses the model's dependence on the language bias of the dataset by changing the answer distribution between the training and test sets. VLind-Bench~\cite{lee2025vlind} introduces a dedicated evaluation framework for language priors in VLMs. Through sequential verification of common-sense knowledge, visual perception, and common-sense bias, it reduces the influence of these confounding factors when assessing language-prior behavior.  ViLP~\cite{luo2024probing} further probes visual-language priors using out-of-distribution question–image–answer triplets and distractor facts embedded in questions, examining whether models rely on textual priors or visual evidence when producing answers.

However, although both Textual Modality Bias and Language Prior may cause the model output to deviate from visual evidence, the sources of information they correspond to are different: the former is reflected in the model's tendency to adopt external information explicitly provided by the input text, while the latter is reflected in the model's prior knowledge or linguistic statistical tendency under specific questioning conditions. Existing research has not yet fully differentiated the role of these two types of factors in cross-modal conflict. Therefore, it is necessary to construct an evaluation framework that can separately characterize the impact of visual information, external text information, and language priors, so as to analyze the decision-making basis of the model in more detail.

\subsection{Multimodal Evaluation Methodology}\label{subsec2.3}

In multimodal model evaluation, discrete outputs and probabilistic information provide two complementary perspectives for analyzing model behavior. Multiple-choice evaluation is widely adopted because it enables explicit comparison among predefined candidate answers. However, its reliability can be affected by selection bias and option position bias. Zheng et al. \cite{zheng2024large} demonstrate that LLMs may exhibit preferences for specific option identifiers, such as A or B, regardless of the semantic content of the corresponding answers. Pezeshkpour and Hruschka \cite{pezeshkpour2024large} further investigate the sensitivity of LLMs to option ordering, showing that rearranging candidate answers can substantially affect model predictions. More recently, Atabuzzaman et al. \cite{atabuzzaman2025benchmarking} extend this investigation to large vision-language models and identify selection biases in multiple-choice visual question answering. These findings highlight the importance of carefully designing option mappings when evaluating model decisions. Therefore, in cross-modal information source attribution tasks, option mappings should be appropriately designed across different evaluation conditions to reduce potential interference from option-related biases.

Beyond the final answer selected by a model, probabilistic information provides additional insights into its preferences among candidate answers. Jiang et al. \cite{jiang2021can} investigate the use of candidate answer probabilities for estimating model confidence in question answering, calculating normalized probabilities to characterize the relative likelihood of different candidate answers. Kumar et al. \cite{kumar2024confidence} further examine the relationship between token-level probabilities and model confidence, deriving internal confidence estimates by extracting and normalizing the probabilities of tokens corresponding to different answer options. These studies demonstrate the potential of token-level probabilistic information for characterizing model preferences beyond the final generated answer.

Probabilistic information has also been explored in uncertainty estimation and hallucination detection. Farquhar et al. \cite{farquhar2024detecting} introduce semantic entropy to quantify uncertainty over the meanings of generated responses by aggregating the probabilities of semantically equivalent outputs. In the vision-language domain, Zhang et al. \cite{zhang2024vl} propose an uncertainty-based framework that estimates the reliability of LVLM responses through semantic clustering of outputs generated from perturbed but semantically equivalent inputs. These studies further demonstrate the value of probabilistic and distributional information in characterizing model behavior and identifying potential hallucinations.

Inspired by these studies, this paper combines two complementary approaches: discrete attribution and soft attribution. Discrete attribution determines the information source corresponding to the final option selected by the model, whereas soft attribution utilizes the log probabilities of Top-K tokens to calculate the relative probability support associated with candidate answers from different information sources. By characterizing model behavior from the perspectives of final answer selection and candidate answer probability support, these two approaches provide complementary criteria for analyzing information source preferences under cross-modal information conflicts.

\section{Problem Formulation and Dateset Construction}\label{sec3}

This paper aims to construct an evaluation benchmark that implements the VTK three-way attribution framework to distinguish the influence of visual information, external textual information, and the model's own prior knowledge on the output of visual language models. To this end, this chapter divides the construction of the benchmark into three stages: Problem Formulation, VTK-Bench Construction, and Data Validation and Quality Control.

\subsection{Problem Formulation}\label{subsec3.1}

This paper focuses on the different information source biases of the VLMs in cross-modal information conflict scenarios. Therefore, we categorize the sources of information that influence the model output into three types: V, which is the actual visual content presented in the input image; T, which is the external information explicitly provided to the model through the input text; and K, which is the prior answer tendency corresponding to the model's existing knowledge and linguistic statistical association, that is, the model may exhibit a tendency to produce a particular answer based on question-relevant knowledge and linguistic associations acquired during pre-training. In this attribution framework, V, T, and K correspond to the preset answers supported by the current image, external text, and prior knowledge, respectively. It should be noted that the K-type answers are not a direct reading of the model's internal knowledge, but rather are validated through the model's performance under specific cue conditions to determine whether they possess the corresponding prior answering tendency.

Based on the above classification of information sources, this paper further distinguishes three phenomena: Visual Factuality (VF), TMB, and LP. VF refers to whether the textual response output by the model is consistent with the objective visual facts in the image. VF refers to the model's ability to correctly extract information from an image based on the question. TMB indicates that when input text information conflicts with visual information, the model tends to adopt information provided by external text rather than faithfully relying on visual evidence in the image. LP refers to the model's tendency to generate answers that conform to common cognition but are inconsistent with current visual evidence when faced with question about visual information, based on its existing knowledge or linguistic statistical associations. While TMB and LP may produce similar erroneous outputs, their underlying information sources differ: the former reflects the model's over-reliance on external textual information, while the latter reflects the influence of the model's prior knowledge on the output. Therefore, simply relying on whether the final output conforms to common perception is insufficient to accurately distinguish between the two phenomena; it is necessary to further attribute the information sources upon which the model output depends.

Addressing the similarities in surface outputs and the differences in information sources between TMB and LP, this paper aims to establish an attribution framework capable of distinguishing the influence of different information sources. The core idea is to construct samples where there are clear conflicts between V, T, and K, making the content supported by these three types of information distinguishable, thus providing a foundation for identifying the model's dependence on different information sources. Through this design, this paper can not only distinguish the model's different tendencies to depend on external textual information and its own prior knowledge, but also further examine whether the model faithfully responds based on current visual evidence. Based on this, this paper uses V, T, and K as a unified attribution object, laying the foundation for subsequent benchmark construction and model behavior analysis under cross-modal information conflict.

To address the similarities in surface outputs and the differences in information sources between TMB and LP, this paper proposes a three-way information source attribution framework based on V, T, and K, aiming to systematically analyze the model's tendency to depend on different information sources under cross-modal information conflict. By constructing conflict scenarios with clear distinctions between the three types of information (V, T, and K), the impact of different information sources on the model output can be examined separately. This allows us to differentiate the model's different dependencies on external textual information and its own prior knowledge, and further evaluate its adherence to current visual evidence. Based on this, this chapter constructs a benchmark around the three types of information sources in VTK, providing a foundation for analyzing the information source bias of VLM and its relationship with multimodal illusions.

\subsection{VTK-Bench Construction}\label{subsec3.2}
\begin{figure*}[t]
    \centering
    \includegraphics[width=\textwidth]{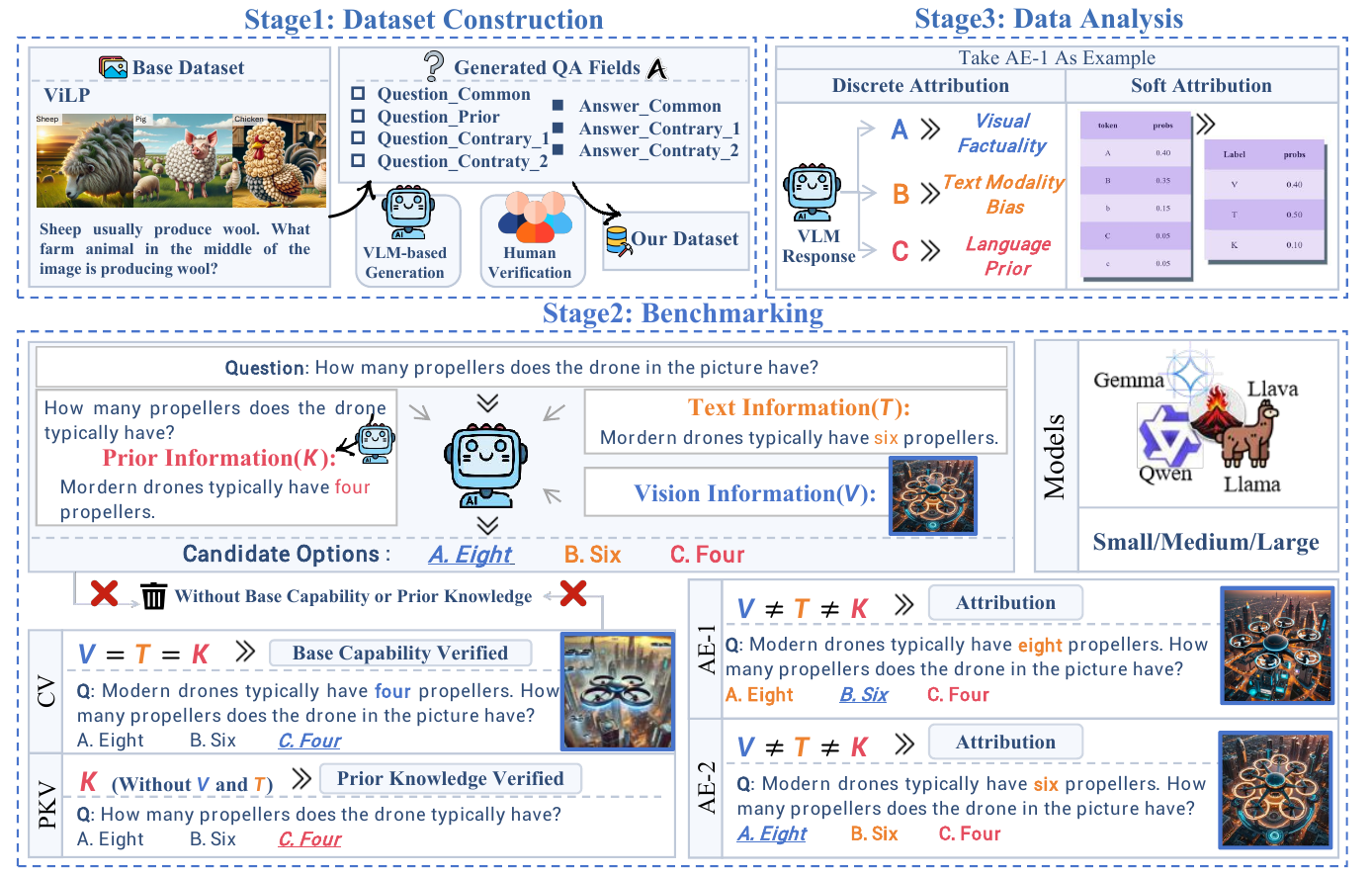}
    \caption{Overview of the VTK-Bench Construction and Evaluation Pipeline.}
    \label{fig:fig_overview}
\end{figure*}

Fig.~\ref{fig:fig_overview} illustrates the overall construction and evaluation pipeline of VTK-Bench, covering dataset construction, benchmark evaluation, and attribution analysis. Based on the aforementioned VTK three-way information source attribution concept, this paper further constructs a benchmark for evaluating the information source bias of VLM. To ensure that the three types of information sources are distinguishable at the sample level, the dataset constructs evaluation samples that satisfy the preset information conflict relationship by combining different questions, images and their corresponding answers.

\subsubsection{Basic Data Sources and Problem Design}\label{subsubsec3.2.1}

In order to construct an evaluation data set that can distinguish the effects of V, T and K, this paper first analyzes the sample coverage of existing visual language evaluation data sets. Although the existing data set contains a certain number of visual language interaction samples, it still has shortcomings in the coverage of abnormal visual scenes and the independent verification of the model's prior knowledge tendencies, making it difficult to directly meet the needs of this article for fine-grained attribution of the three types of information sources. Therefore, this article selects ViLP ~\cite{luo2024probing} as the basic data source, and performs targeted sample addition and structural processing on the basis of retaining its original images and related problem information. Specifically, ViLP's original question and answer pairs are mainly used to determine whether the model is misled by K. Its evaluation goal focuses on binary judgments and lacks the multiple sets of questions and answers required to support the three-way mutually exclusive attribution of V, T and K. It also lacks related fields for independent verification of the model's prior knowledge tendency. Therefore, it is necessary to further expand the question types and sample structure based on the original data.

Focusing on the three-way attribution requirements of VTK, this article constructs three types of questions: Common-sense Question, Anomalous Question and Prior Question. Both common-sense questions and anomalous questions consist of two parts: a text prompt containing preset answer information, and a question about the image content. The prior question is used to detect whether the model shows the preset prior knowledge tendency of the sample without providing target images and external text information. Its expected answer is consistent with the preset common-sense answer in the sample.

\subsubsection{Sample Construction and Cross-matching}\label{subsubsec3.2.2}

In the process of supplementing the basic data, this article follows the semantic consistency constraints between questions, images and answers, and uses the image and answer information in the original samples to generate common-sense questions and two types of anomalous questions. Specifically, the information contained in different answers is integrated into the original questions respectively to form common-sense questions and anomalous questions with corresponding text prompts, that is, T, and ensure that each question has a clear expected answer under its corresponding image. For a prior questions, this article extracts the target subject and attribute information based on the original question, and uses a large language model to generate targeted a prior questions, so that it can detect whether the model tends to give preset common sense answers without providing V and T.

To achieve VTK three-way information source identification and attribution, this paper further employs a non-homogeneous cross-pairing strategy, combining questions and images from different sources so that V, T, and K each point to different preset answers. This design is crucial for distinguishing the three information sources: if questions and images are always paired according to their original correspondence, V and T may point to the same answer, making it difficult for the model output to differentiate whether it is primarily influenced by textual or visual information. Through cross-pairing, this paper ensures that the preset answers corresponding to the three types of information sources are always mutually exclusive in the core evaluation, thus providing a clear basis for subsequent attribution.

\subsubsection{Data Representation and Structuring}\label{subsubsec3.2.3}
In order to facilitate data management and subsequent evaluation, this article uses a unified JSON format to organize each sample. Each sample contains 11 fields, covering four types of information: sample identification, questions, answers and images. The specific field definitions are shown in TABLE~\ref{tab:table_JSON}. Throughout the paper, the terms common-sense and anomalous are used to describe the two types of information, with anomalous corresponding to the original contrary fields in the JSON data.
\begin{table*}[t]
    \centering
        \caption{Description of the JSON Fields in Each Sample. }
        \label{tab:table_JSON}
        \resizebox{\textwidth}{!}{
        \begin{tabular}{C{0.15\textwidth}C{0.35\textwidth}C{0.5\textwidth}}
            \toprule
            Notation & JSON Field & Description\\
            \midrule
            \(\mathrm{ID}\) & id & Unique sample identifier\\
            \(Q_{\mathrm{CS}}\) & question\_common & Common-sense question\\
            \(Q_{\mathrm{PR}}\) & question\_prior & Prior question\\
            \(Q_{\mathrm{AN1}}\) & question\_contrary\_1 & Anomalous question 1\\
            \(Q_{\mathrm{AN2}}\) & question\_contrary\_2 & Anomalous question 2\\
            \(A_{\mathrm{CS}}\) & answer\_common & Answer to the common-sense question\\
            \(A_{\mathrm{AN1}}\) & answer\_contrary\_1 & Answer to anomalous question 1\\
            \(A_{\mathrm{AN2}}\) & answer\_contrary\_2 & Answer to anomalous question 2\\
            \(I_{\mathrm{CS}}\) & image\_common & Image for the common-sense question\\
            \(I_{\mathrm{AN1}}\) & image\_contrary\_1 & Image for anomalous question 1\\
            \(I_{\mathrm{AN2}}\) & image\_contrary\_2 & Image for anomalous question 2\\
            \bottomrule
        \end{tabular}
        }
\end{table*}

\begin{figure*}[t]
    \centering
    \includegraphics[width=\textwidth]{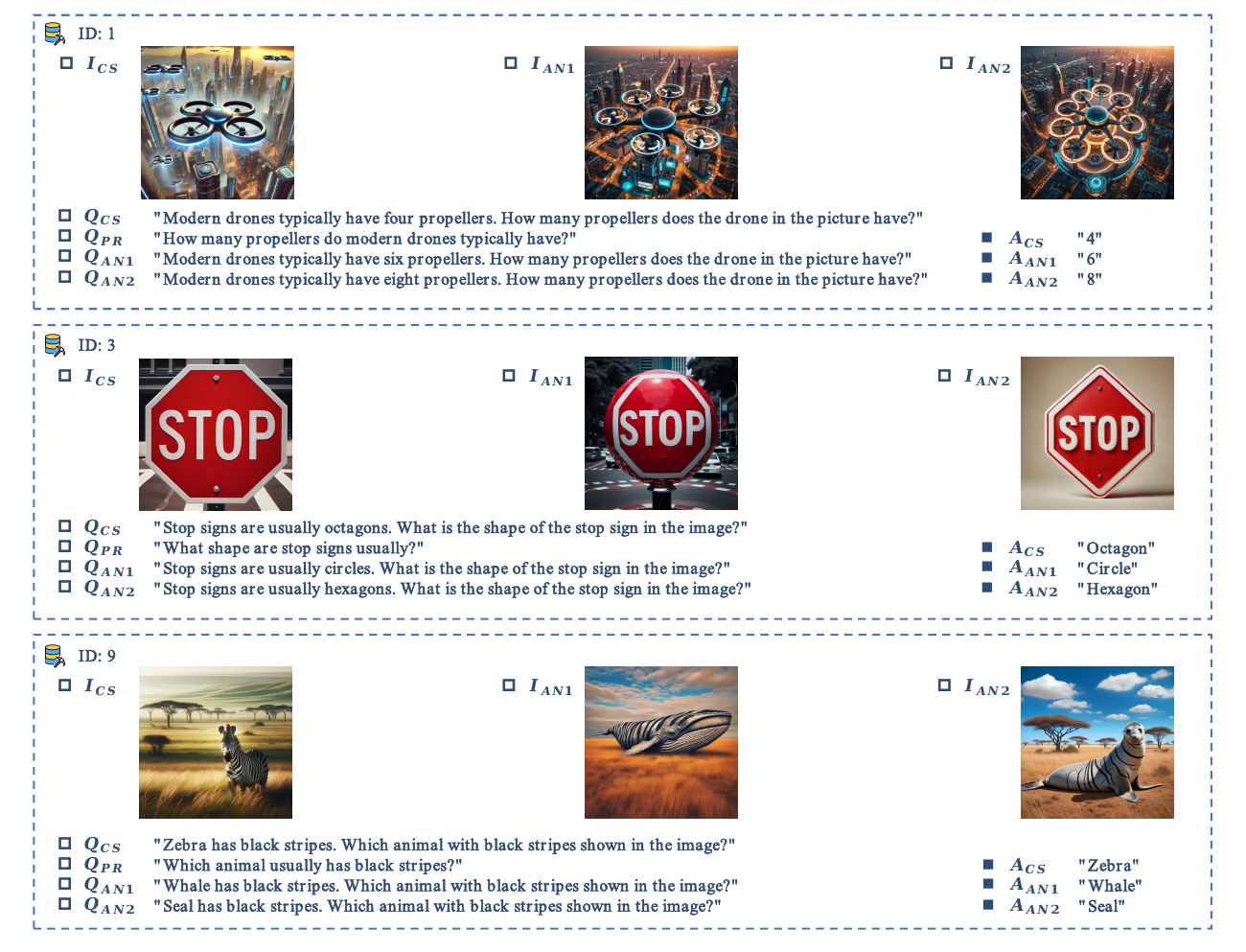}
    \caption{Examples of Constructed Benchmark Samples.}
    \label{fig:fig_sample}
\end{figure*}

To further illustrate the constructed data structure, Fig.~\ref{fig:fig_sample} presents three representative samples from VTK-Bench.

Among them, \(Q_{\mathrm{PR}}\) is used to independently detect the prior knowledge tendency of the model, and its expected answer is consistent with \(A_{\mathrm{CS}}\); the remaining question, answer and image path fields organize the original sample information through corresponding relationships, and are combined according to the preset pairing rules in the core evaluation stage. In order to ensure the integrity and consistency of the data structure, automatic verification is performed after data is written, including field integrity, image path validity, question and answer mapping relationship, candidate answer mutual exclusivity, duplicate sample detection and JSON format verification. Data that passes the verification enters the subsequent verification and quality control process.

\subsection{Data Validation and Quality Control}\label{subsec3.3}

\subsubsection{Solvability Verification}\label{subsubsec3.3.1}

To ensure that the constructed dataset meets the basic requirements for subsequent attribution evaluation, this paper validates the automatically generated samples from two aspects: basic answering ability and prior answering tendency. The former is used to check whether the model has the basic ability to complete the target task when the image, question, and auxiliary text information are consistent; specifically, by inputting a normal question and its corresponding image, it checks whether the model output is consistent with the preset answer. The latter is used to check whether the model exhibits a pre-set tendency to answer based on the sample; specifically, without providing the target image and external text information, only a pre-set question is input, and it checks whether the model output is consistent with the preset pre-set answer. This paper uses Qwen3-VL-$8$B-Instruct as the data quality detection model, performs the two types of verifications mentioned above, and records the model's response and its matching results with the preset answer.

\subsubsection{Failure Analysis and Processing}\label{subsubsec3.3.2}

Based on the verification results, this paper categorizes the samples that failed verification into two types. The first type is data construction errors, where the automatically generated question does not match the attributes of the image subject or target. For example, the image may actually show the number of propellers on a drone, but the generated question asks about the drone's typical flight location, resulting in a lack of correspondence between the question and the target's visual attributes. For this type of sample, this paper manually verifies the source of the question and corrects the question or related data to restore the semantic consistency between the question, image, and preset answer.

The second category is model validation failure, where the sample itself meets the pre-set design requirements, but the probe model fails the corresponding validation. This may stem from insufficient basic response capabilities of the model or failure to demonstrate the pre-set prior response tendency of the sample, and does not necessarily mean that the sample construction is flawed. Therefore, this paper retains such samples and their validation records instead of directly classifying them as invalid data. Considering that different models do not possess entirely the same prior knowledge, the failure of a single probe model to demonstrate the pre-set prior response tendency is not sufficient to negate the validity of the sample itself. In the subsequent formal evaluation, this paper will select samples that meet the corresponding attribution premises based on the validation results of each tested model, and record the number of valid samples and their proportion for each model.

For samples confirmed to have data construction errors after manual review, this paper corrects the relevant issues or data and re-executes the above verification to check the effect of the correction and whether it introduces new problems. This correction and retesting process continues until there are no unresolved data construction errors. It should be noted that the verification results during the dataset construction phase are mainly used to assist in quality checks and are not the sole basis for sample retention or rejection. For samples with reasonable design but that fail model validation, this paper retains them.

\section{Experiments}\label{sec4}

\subsection{VTK-Bench Evaluation Protocol}\label{subsec4.1}

This paper adopts a four-stage evaluation protocol consisting of two basic verification stages and two core evaluation stages. The first two stages are used to evaluate the model's basic response ability and its tendency to respond to pre-set answers, providing necessary prerequisites for subsequent attribution analysis. The latter two stages construct conflict scenarios between visual information, external textual information, and prior knowledge corresponding to different candidate answers, combining the same sample into two test data sets to examine the model's selection preference among different information sources. The specific design of each stage is as follows.

Consistency Verification (CV): Basic Ability Validation. This stage inputs a common sense question and its corresponding common sense image, requiring the model to select the common sense answer from the candidate options. As shown in Eq.~\eqref{eq.eq_cv}, the preset answers corresponding to V, T, and K must be consistent. Therefore, this stage mainly tests whether the model can correctly complete the target task under the condition that the three types of information are consistent. Samples that fail this stage will not be included in subsequent analysis..
\begin{align}
    y^V=y^T=y^K 
    \label{eq.eq_cv}
\end{align}
where the preset answers corresponding to the three types of information sources are \(y^V\), \(y^T\), and \(y^K\), respectively.

Prior knowledge existence verification (PKV): This step only inputs a prior question, without providing a target image, and requires the model to select the option consistent with the common-sense answer. Since this step does not provide visual evidence or external textual cues for the current sample, its purpose is to verify whether the model possesses prior knowledge tendencies consistent with the preset common-sense answer. Samples that fail this step will not proceed to subsequent analysis. The reason for this step is that if the model itself does not possess K, its output of common-sense answers in the core evaluation cannot be directly attributed to LP, but may simply be a chance guess.

Attribution Evaluation (AE-1 and AE-2): Cross-modal Conflict Diagnosis. Samples that pass both CV and PKV enter the AE stages. In these stages, anomalous questions and anomalous images from different anomalous groups within the same sample are cross-paired to construct cross-modal conflict scenarios, as illustrated in Eq.~\eqref{eq.eq_ae}. Specifically, the anomalous question from one group is paired with the anomalous image from the other group, thereby establishing conflicts among V, T, and K. These three information sources are associated with three mutually exclusive pre-defined answers, respectively, enabling their competing influences on the model's decision to be separately examined. The model is required to select one of the three candidate options. AE-1 and AE-2 differ in the option-to-source mappings of V and T, while K remains mapped to the same option. Based on the selected option and its corresponding information source, the model's decision is classified into one of three categories: VF, TMB, or LP.
\begin{align}
    {y^V}_{i,r}\neq {y^T}_{i,r},\ {y^V}_{i,r}\neq {y^K}_{i,r},\ {y^T}_{i,r}\neq {y^K}_{i,r}
    \label{eq.eq_ae}
\end{align}
where \(i\) represents the original sample index, and \(r\in\{1,\ 2\}\) means that the two test data corresponding to this sample match AE-$1$ and AE-$2$ respectively. It should be noted that in VTK-Bench, one set of samples corresponds to two sets of test data.

AE-$1$ and AE-$2$ use the same three types of information content, but employ different combinations of questions and images, and set different options and semantic role mapping relationships. Therefore, the two stages constitute two test datasets, used to examine the model's information source selection preferences under different input combinations and option mapping conditions. The attribution results of the two stages are recorded separately and used together for subsequent statistical analysis.

To avoid a fixed correspondence between candidate option positions and semantic roles, this paper employs an option role substitution strategy. The composition of the three candidate answers remains consistent across all evaluation stages; that is, A, B, and C correspond to \(A_{\mathrm{CS}}\), \(A_{\mathrm{AN1}}\), and \(A_{\mathrm{AN2}}\), respectively. However, the semantic roles represented by each option change with each evaluation stage. Therefore, option A represents the pre-defined correct answer in the basic validation stage, while representing the answer corresponding to the linguistic prior in AE stage. This substitution design reduces the model's dependence on fixed option positions while maintaining the unchanged composition of candidate options. The correspondence between each evaluation stage and the dataset fields is shown in Table~\ref{tab:table_map}.

\begin{table*}[t]
    \centering
    \caption{Dataset Field and Option-to-Source Mappings across Evaluation Stages.}
    \label{tab:table_map}    
    \resizebox{\textwidth}{!}{
        \begin{tabular}{lcccC{0.23\textwidth}C{0.2\textwidth}}
            \toprule
            Stage & Question & Expected answer & Image & Option-to-source mapping & Relation among V, T, and K\\
            \midrule
            CV & \(Q_{\mathrm{CS}}\) & \(A_{\mathrm{CS}}\) & \(I_{\mathrm{CS}}\) & A$\to$ T, V, K & T = V = K\\
            PKV & \(Q_{\mathrm{PR}}\) & \(A_{\mathrm{CS}}\) & - & A$\to$K & K only\\
            AE-$1$ & \(Q_{\mathrm{AN1}}\) & \(A_{\mathrm{AN2}}\) & \(I_{\mathrm{AN2}}\) & A$\to$K, B$\to$T, C$\to$V & T $\neq$ V $\neq$ K\\
            AE-$2$ & \(Q_{\mathrm{AN2}}\) & \(A_{\mathrm{AN1}}\) & \(I_{\mathrm{AN1}}\) & A$\to$K, B$\to$V, C$\to$T & T $\neq$ V $\neq$ K\\
            \bottomrule
        \end{tabular}
    }
\end{table*}

For AE data that passed the basic validation, this paper uses two methods to analyze its information sources from two levels: the final selection of the model and the probability distribution of candidate tokens.

Discrete Attribution (DA): Based on the response generated by the model, the final candidate answer is identified and attributed to its corresponding information source category. Let \(\hat{y}_{i,r}\) denote the corresponding model response. The response is semantically matched against the preset candidate answers to identify the selected answer, based on which the attribution is determined according to the predefined mapping rules:
\begin{align}
    A(\hat{y}_{i,r}) =
    \begin{cases}
    V, & {Match}(\hat{y}_{i,r}, y_{i,r}^{V}) = 1, \\
    T, & {Match}(\hat{y}_{i,r}, y_{i,r}^{T}) = 1, \\
    K, & {Match}(\hat{y}_{i,r}, y_{i,r}^{K}) = 1.
    \end{cases}
    \label{eq.eq_DA}
\end{align}
where \({Match}(\ ,\ )\) represents the semantic matching function.

Soft Attribution (SA): We further utilize the Top-K Tokens output by the model and their corresponding Logprobs to measure the relative probability support of different information sources for the model output. First, we calculate the corresponding probability based on the Token's Logprob:
\begin{align}
    p_{i,r}(t)=exp(l_{i,r}(t))
    \label{eq.eq_SA1}
\end{align}
where \(t\) represents a candidate token in the Top-K Token, and \(l_{i,r}(t)\) represents the Logprob corresponding to Token \(t\) original sample.

Considering that the same preset option may correspond to different token representations, this paper, based on semantic matching rules, merges tokens in the Top-K Tokens that are semantically equivalent to the same preset option, and sums their corresponding probabilities to obtain the probability support value of each information source category in each inference:
\begin{align}
P_{i,r}^{c}
&= \sum_{t \in G_{i,r}^{c}} p_{i,r}(t),
\quad c \in \{V,T,K\}.
\label{eq.eq_SA2}
\end{align}
where \(G_{i,r}^{c}\) represents the set of tokens that semantically match the preset option corresponding to information source category \(c\), and \(P_{i,r}^c\) represents the probability support value of category \(c\) in this inference.

Subsequently, to measure the relative probability support of different information sources in each AE test data point, this paper normalizes the probability support values of the three information sources (V, T, and K) in each inference, obtaining the soft attribution score for the corresponding test data:
\begin{align}
S_{i,r}^{c}
&= \frac{P_{i,r}^{c}}{\sum_{d \in \{V,T,K\}} P_{i,r}^{(d)}},
\quad c \in \{V,T,K\}.
\label{eq.eq_SA3}
\end{align}

\subsection{Experimental Setup}\label{subsec4.2}

\subsubsection{Evaluation Model and Experimental Environment}\label{subsubsec4.2.1}
This paper selects $11$ mainstream open-source VLMs for cross-modal evaluation. All models are called through a unified OpenAI-compatible interface to ensure the consistency of the evaluation protocol between different models. The model sizes cover $2$B to $32$B parameter counts, encompassing dense architectures and mixture-of-experts architectures, and covering as many different model families as possible to examine the impact of model size and architectural differences on cross-modal conflict handling behavior. Specific models include: gemma-$4$-$31$B-it \cite{gemmateam2026gemma4}, Qwen3.8-$27$B \cite{qwen38}, Qwen3-VL ($2$B, $4$B, $8$B, $32$B and $30$B-A$3$B) \cite{qwen3vltechnicalreport}, Qwen2.5-VL ($7$B and $32$B) \cite{Qwen2.5-VL}, llava-onevision-qwen2-$7$b-ov \cite{li2024llavaonevisioneasyvisualtask} and llama3-llava-next-$8$b-hf \cite{li2024llavanext-strong}.

In terms of inference settings, all models use a unified prompt word template. The system prompts the model to select one of three candidate options A, B and C, as the answer, and outputs only the corresponding uppercase letter, without any explanatory text, punctuation, or other content. Simultaneously, the system obtains the candidate tokens and their corresponding logprobs at the model's output position through a unified interface, providing probabilistic information for subsequent SA analysis. The inference parameters are set as follows: temperature = $0$, top\_logprobs = $5$.

\subsubsection{Evaluation Indicators}\label{subsubsec4.2.2}

This paper reports on the model performance from the following three aspects.

Verification Metrics (Verif.): The proportion of samples that pass both the CV and PKV sub-stages is statistically analyzed to measure the sample coverage of the model in meeting the preconditions for AE. Specifically, only an original sample that passes both CV and PKV is included in the subsequent AE analysis. This indicator reflects the effective sample coverage of the model on the current benchmark task, providing necessary preliminary reference for cross-sectional comparisons of attribution results from different models.

Discrete Attribution Score (DA-score): Based on the responses generated by the models in AE-$1$ and AE-$2$, the final candidate options are determined, and each decision is categorized into three classes—VF, TMB, or LP—according to the corresponding option role mapping relationships. Using valid test data as the statistical unit, the DA-score is defined as:
\begin{align}
DA^{c}
&= \frac{1}{N} \sum_{i=1}^{M} \sum_{r=1}^{2}
\mathbb{I}(D_{i,r} = c),
\quad c \in \{\mathrm{VF}, \mathrm{TMB}, \mathrm{LP}\}
\label{eq.eq_DAs}
\end{align}
where \(D_{i,r}\) represents the discrete attribution result that corresponds to the valid test data matching \(A(\hat{y}_{i,r})\).

Soft Attribution Score (SA-score): To further measure the relative probability support of different information sources in the model output, this paper uses the Top-K Tokens returned by the model output position and their corresponding Logprobs to calculate the soft attribution score. First, the probability of each Token is calculated based on its Logprob, and according to a predefined option role mapping relationship, the probabilities of semantically matched Tokens are categorized into three information sources: V, T, and K. Then, the normalized probability support score for each of the three information sources is calculated for each AE test data point, and the average score of all valid AE test data points is calculated at the dataset level. The SA-score is defined as:
\begin{align}
SA^{c}
&= \frac{1}{N} \sum_{i=1}^{M} \sum_{r=1}^{2}
S_{i,r}^{c},
\quad c \in \{V,T,K\}
\label{eq.eq_SAs}
\end{align}
where \(S_{i,r}^c\) represents the normalized probability support score, and its specific calculation is shown in Eq.~\eqref{eq.eq_SA3}. \({SA}^c\) reflects the model's preference for information from different sources.

According to the above definition, for the DA-score and SA-score indices, the sum of the three soft attribution scores is $1$, that is:
\begin{align}
\sum_{c \in \{VF,TMB,LP\}} DA^{c} &= 1, \notag \\
\sum_{c \in \{V,T,K\}} SA^{c} &= 1.
\label{eq.eq_DA_SA}
\end{align}
DA-score and SA-score characterize the model's dependence on different information sources from two perspectives: final choice and relative probability support, respectively. The former reflects the attribution distribution of the model's final decision, while the latter further characterizes the relative support of different information sources in the probability quality of Top-K Tokens. These two metrics complement each other, helping to analyze other probabilistic support information that might be overlooked by relying solely on discrete statistics based on the final option, and providing different measurement perspectives for analyzing the model's behavior under cross-modal information conflict conditions.

\subsection{TMB and LP Analysis}\label{subsec4.3}

\begin{table*}[t]
    \centering
    \caption{Discrete attribution results of 11 open-source VLMs. The reported VF, TMB, and LP values correspond to \(DA^{\mathrm{VF}}\), \(DA^{\mathrm{TMB}}\), and \(DA^{\mathrm{LP}}\), respectively. Models are sorted in descending order of visual fidelity.}
    \label{tab:tab_DA}
    \resizebox{\textwidth}{!}{%
        \begin{tabular}{C{0.45\textwidth}C{0.2\textwidth}ccc}
            \toprule
            \textbf{VLM} & \textbf{Verif.} & \textbf{VF} $\uparrow$ & \textbf{TMB} $\downarrow$  & \textbf{LP} $\downarrow$ \\
            \midrule
            gemma-4-$31$B-it               & $93.7$\% & $79.7$\%  & $2.1$\%  & $18.2$\% \\
            Qwen3.8-$27$B                  & $94.3$\% & $69.9$\%  & $1.8$\%  & $28.3$\% \\
            Qwen3-VL-$30$B-A$3$B-Instruct    & $96.3$\% & $69.7$\%  & $1.9$\%  & $28.4$\% \\
            Qwen3-VL-$32$B-Instruct        & $97.0$\% & $63.9$\%  & $6.5$\%  & $29.6$\% \\
            Qwen3-VL-$8$B-Instruct         & $95.7$\% & $55.7$\%  & $10.5$\% & $33.8$\% \\
            Qwen2.5-VL-$32$B-Instruct      & $97.0$\% & $51.2$\%  & $8.9$\%  & $39.9$\% \\
            llava-onevision-qwen2-$7$b-ov  & $98.0$\% & $43.9$\%  & $8.3$\%  & $47.8$\% \\
            llama3-llava-next-$8$b-hf      & $93.7$\% & $42.5$\%  & $9.6$\%  & $47.9$\% \\
            Qwen3-VL-$4$B-Instruct         & $95.3$\% & $37.6$\%  & $14.7$\% & $47.7$\% \\
            Qwen2.5-VL-$7$B-Instruct       & $95.0$\% & $34.2$\%  & $13.7$\% & $52.1$\% \\
            Qwen3-VL-$2$B-Instruct         & $95.7$\% & $17.9$\%  & $39.6$\% & $42.5$\% \\
            \bottomrule
        \end{tabular}%
    }
\end{table*}

\subsubsection{Overall Distribution}\label{subsubsec4.3.1}

As shown in Table~\ref{tab:tab_DA}, among the $11$ models evaluated, LP was the most prevalent failure type, with $10$ models having an LP percentage more than $10$\% higher than TMB. The LP percentages of each model ranged from $18.2$\% to $52.1$\%, with even gemma-4-$31$B-it, which performed well in visual attribution, still having an LP percentage of $18.2$\%.

In contrast, the overall TMB percentage ranged from $1.8$\% to $39.6$\%, which is relatively low among most models. Notably, Qwen3-VL-$2$B-Instruct's TMB and LP percentages were $39.6$\% and $42.5$\%, respectively, which are quite close, indicating that this model exhibits a strong reliance on external textual information and prior biases related to the question in cross-modal conflict scenarios.

The above results indicate that LP and TMB are two failure modes that need to be distinguished. Attributing both to LP may misjudge the model's dependence on external textual information as its own prior bias, thus affecting the judgment of the source of multimodal hallucinations.

\subsubsection{Failure Patterns Across Models}\label{subsubsec4.3.2}

As shown in Table~\ref{tab:tab_DA}, different models exhibit differences in their attribution distributions across the three information sources: V, T, and K. For example, the VF proportions of Qwen2.5-VL-$7$B-Instruct, Qwen3-VL-$8$B-Instruct, llama3-llava-next-$8$b-hf, and llava-onevision-qwen2-$7$b-ov are $34.2$\%, $37.6$\%, $42.5$\%, and $43.9$\%, respectively; the corresponding TMB proportions are $13.7$\%, $10.5$\%, $9.6$\%, and $8.3$\%; and the corresponding LP proportions are $52.1$\%, $33.8$\%, $47.9$\%, and $47.8$\%. These models differ by $9.7$\% in VF proportions, $5.4$\% in TMB proportions, and $18.3$\% in LP proportions, indicating that even models of similar size may exhibit different information source preferences.

Among models with a high proportion of VF, gemma-4-$31$B-it exhibits VF, TMB, and LP attribution ratios of $79.7$\%, $2.1$\%, and $18.2$\%, respectively, demonstrating high VF and low TMB. However, its LP attribution still accounts for a certain proportion, indicating that a high VF proportion does not necessarily mean that prior biases related to the question have completely disappeared.

\subsubsection{Case Studies}\label{subsubsec4.3.3}

The representative cases shown in the lower-right part of Fig.~\ref{fig:fig_overview} illustrate typical model responses corresponding to the three information-source attribution categories. All three cases are based on Sample ID 1, which concerns the number of objects in the image and provides three candidate answers: A (4), B (6), and C (8).
Case $1$ – VF. gemma-4-$31$B-it selects C in AE-$1$ and B in AE-$2$. According to the corresponding option-to-source mappings, both selections correspond to V, indicating that the model consistently relies on visual information in both evaluation stages. Accordingly, both responses are attributed to VF.

Case $2$ – TMB. Qwen3-VL-$2$B-Instruct selects B in AE-$1$ and C in AE-$2$. According to the respective option-to-source mappings, both selections correspond to T, indicating that the model consistently relies on external textual information rather than visual evidence. Accordingly, both responses are attributed to TMB.

Case $3$ – LP. Qwen3-VL-$4$B-Instruct selects A in AE-$2$, where A corresponds to K. Thus, its response is attributed to LP. Notably, the model does not select A in AE-1, indicating that the LP attribution is observed only in AE-2 for this sample.

The above cases demonstrate that when faced with the same cross-modal conflict content, different models may exhibit different information source biases, and also illustrate that it is insufficient to judge the basis of a model's answer solely based on whether the final answer conforms to common sense.

\subsection{Attribution across Model Variants}\label{subsec4.4}

\subsubsection{Scaling Trends}\label{subsubsec4.4.1}

\begin{figure}[t]
    \centering
    \includegraphics[width=\textwidth]{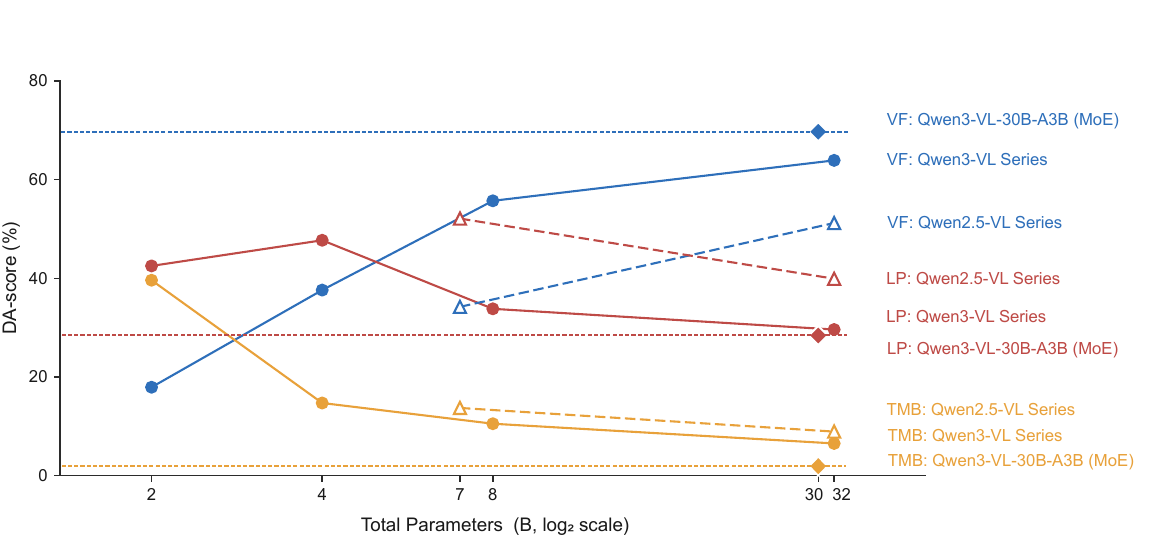}
    \caption{Discrete attribution shares across model scales and generations. The three colors represent VF, TMB, and LP, respectively. Solid lines with circular markers denote Qwen3-VL models, dashed lines with triangular markers denote Qwen2.5-VL models, and diamond markers denote the Qwen3-VL-$30$B-A$3$B MoE model. The horizontal axis represents total model parameters on a base-$2$ logarithmic scale. Horizontal dashed segments around the MoE markers are included solely to facilitate comparison with the Qwen3-VL-$32$B dense model and do not represent error bars or confidence intervals.}
    \label{fig:fig_qwen}
\end{figure}

As shown in Fig.~\ref{fig:fig_qwen}, as the size of the Qwen3-VL series models increased from $2$B to $32$B, different attribution categories exhibited different trends. Specifically, the proportion of VF increased from $17.9$\% to $37.6$\%, $55.7$\%, and $63.9$\%, while the proportion of TMB gradually decreased from $39.6$\% to $14.7$\%, $10.5$\%, and $7.1$\%. This result indicates that in this series of models, increased model size is accompanied by a higher proportion of VF and a lower proportion of TMB.	

Unlike VF and TMB, LP did not decrease monotonically with increasing model size. The LP proportions for Qwen3-VL-$2$B-Instruct, Qwen3-VL-$4$B-Instruct, and Qwen3-VL-$8$B-Instruct were $42.5$\%, $47.7$\%, and $33.8$\%, respectively, showing a trend of first increasing and then decreasing; in the $32$B model, this proportion further decreased to $29.6$\%. This phenomenon indicates that an increase in the proportion of VF does not necessarily lead to a synchronous weakening of the impact of LP. Changes in LP may also be influenced by factors such as model training data, pre-training knowledge, and instruction fine-tuning.

\subsubsection{Generation and Architecture}\label{subsubsec4.4.2}

Besides model size, model generations and architectures may also be related to the distribution of information source attribution. This paper compares the experimental results of the Qwen2.5-VL and Qwen3-VL series to examine the performance differences of models with different generations and architectures at similar scales.

At the $32$B scale, the proportions of VF, TMB, and LP for Qwen2.5-VL-$32$B-Instruct are $51.2$\%, $8.9$\%, and $39.9$\%, respectively, while the proportions for visual attribution for Qwen3-VL-$32$B-Instruct are $63.9$\%, $6.5$\%, and $29.6$\%. In smaller-scale models, the proportions of VF, TMB, and LP for Qwen2.5-VL-$7$B-Instruct are $34.2$\%, $13.7$\%, and $52.1$\%, respectively; within the Qwen3-VL series, the corresponding proportions for Qwen3-VL-$4$B-Instruct are $37.6$\%, $14.7$\%, and $47.7$\%, respectively, while those for Qwen3-VL-$8$B-Instruct are $55.7$\%, $10.5$\%, and $33.8$\%, respectively. These results indicate that, under the test settings in this paper, newer generation models within the same model series exhibit more favorable attribution distribution characteristics.

In the MOE architecture, the proportions of VF, TMB, and LP in Qwen3-VL-$30$B-A$3$B-Instruct, which has a similar number of parameters to $32$B, are $69.7$\%, $1.9$\%, and $28.4$\%, respectively. This result indicates that, under the test settings in this paper, the MoE architecture model exhibits superior attribution characteristics compared to traditional dense architectures in this type of task.

Overall, the experimental results show differences in attribution across different generations, architectures, and sizes of models for the three types of information sources. Model size may be related to improved visual information utilization, but changes in LP did not exhibit the same monotonic trend. This also indicates that measuring a model's performance in cross-modal conflict scenarios solely by model size or a single performance metric is insufficient; it is also necessary to examine visual information, external textual information, and prior biases related to the question separately.
\begin{table*}[t]
    \centering
    \caption{Soft attribution results of 11 open-source VLMs. The reported V, T, and K values correspond to \(SA^{V}\), \(SA^{T}\), and \(SA^{K}\), respectively. Models are sorted according to their order in Table~\ref{tab:tab_DA})}
    \label{tab:tab_SA}
    \resizebox{\textwidth}{!}{%
        \begin{tabular}{C{0.55\textwidth}C{0.12\textwidth}C{0.12\textwidth}C{0.12\textwidth}}
            \toprule
            \textbf{VLM} & \textbf{V} $\uparrow$ & \textbf{T} $\downarrow$ & \textbf{K} $\downarrow$ \\
            \midrule
            gemma-4-$31$B-it               & $79.5$\% & $2.1$\%  & $18.4$\% \\
            Qwen3.8-$27$B                  & $66.3$\% & $5.9$\%  & $27.8$\% \\
            Qwen3-VL-$30$B-A$3$B-Instruct  & $70.1$\% & $1.7$\%  & $28.2$\% \\
            Qwen3-VL-$32$B-Instruct        & $63.2$\% & $7.1$\%  & $29.7$\% \\
            Qwen3-VL-$8$B-Instruct         & $55.2$\% & $11.2$\% & $33.6$\% \\
            Qwen2.5-VL-$32$B-Instruct      & $48.3$\% & $12.9$\% & $38.8$\% \\
            llava-onevision-qwen2-$7$b-ov  & $41.6$\% & $11.7$\% & $46.7$\% \\
            llama3-llava-next-$8$b-hf      & $41.3$\% & $12.1$\% & $46.6$\% \\
            Qwen3-VL-$4$B-Instruct         & $37.7$\% & $14.8$\% & $47.5$\% \\
            Qwen2.5-VL-$7$B-Instruct       & $33.4$\% & $17.8$\% & $48.8$\% \\
            Qwen3-VL-$2$B-Instruct         & $18.8$\% & $38.5$\% & $42.7$\% \\
            \bottomrule
        \end{tabular}%
    }
\end{table*}

\subsection{Analysis of DA and SA}\label{subsec4.5}

\subsubsection{Attribution Consistency}\label{subsubsec4.5.1}

To assess the consistency between the two attribution methods, this paper compares the model rankings obtained from DA-score and SA-score (Tables~\ref{tab:tab_DA} and \ref{tab:tab_SA}) across the three information sources: V, T, and K. Here, V, T, and K correspond to VF, TMB, and LP in DA-score, respectively. Spearman's rank correlation coefficient is employed to measure the consistency between the relative model rankings:
\begin{align}
    \rho_c=Corr\left(R^{c}_{\mathrm{Discrete}},R^{c}_{\mathrm{Soft}}\right),c \in \{V,T,K\} \label{eq.eq_sort}
\end{align}
where \(c\) represents the information source category, \(R^{c}_{\mathrm{Discrete}}\) and \(R^{c}_{\mathrm{Soft}}\) represent the model rank vectors obtained according to DA-score and SA-score, respectively. By calculating the Pearson correlation coefficient between the two sets of rank vectors, the Spearman rank correlation coefficient \(\rho_c\) corresponding to \(c\) is obtained, which takes the value range of \([-1, 1]\). The closer the value is to $1$, the more consistent the model rankings obtained by the two attribution methods are.

Experimental results show that the \(\rho_c\) values for the three attribution categories are $0.991$, $0.900$, and $0.964$, respectively, indicating that the two methods have high consistency in the relative ranking of the models. Among them, category V has the highest correlation, while category T has a relatively lower correlation.

In terms of specific numerical values, the average absolute biases of the V, T, and K attribution categories are $1.25$\%, $1.89$\%, and $0.77$\%, respectively, with maximum biases of $3.62$\%, $4.19$\%, and $3.35$\%. Overall, the attribution results obtained by the two methods are quite similar, but differences still exist in some models and categories.

\subsubsection{Attribution Differences}\label{subsubsec4.5.2}

Although DA and SA exhibit strong consistency in their relative model rankings, differences remain in their numerical attribution scores. To quantify these differences, the attribution difference for each information source is defined as:
\begin{align}
    \Delta{c}=S^{c}_{\mathrm{Soft}}-S^{c}_{\mathrm{Discrete}},c \in \{V,T,K\}\label{eq.eq_delta}
\end{align}

A positive value indicates that SA receives a higher proportion of support for the corresponding category than DA.
\begin{figure*}[t]
    \centering
    \includegraphics[width=\textwidth]{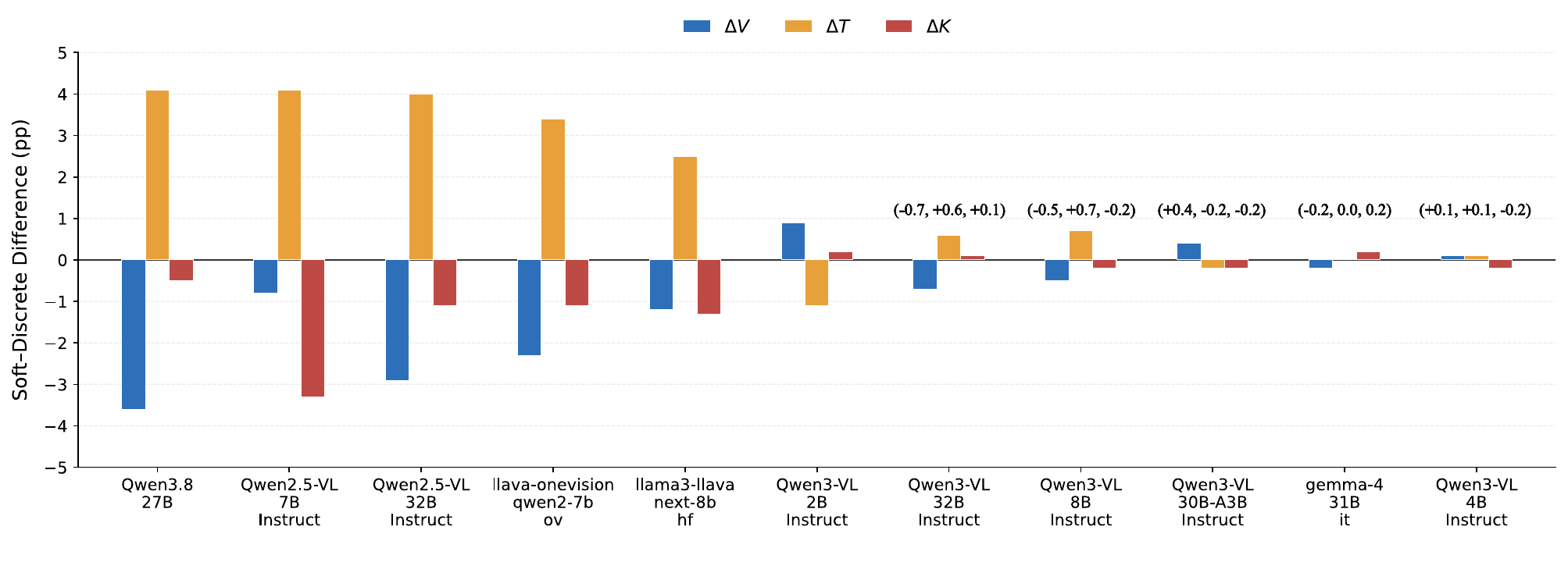}
    \caption{Differences between soft and discrete attribution results across models. $\Delta V$, $\Delta T$, and $\Delta K$ denote the differences between soft attribution scores and discrete attribution proportions for visual information, textual information, and prior knowledge, respectively. Positive and negative values indicate higher and lower proportions under soft attribution, respectively. Models are ordered by the descending range of their three attribution differences. Numerical annotations report the corresponding $(\Delta V, \Delta T, \Delta K)$ values for models whose three differences are all within $1$ pp of zero.}
    \label{fig:fig_bar}
\end{figure*}

Of the signed mean differences across the $11$ models, \(\Delta{V}\) is $-1.0$\%, \(\Delta{T}\) is $+1.7$\%, and \(\Delta{K}\) is $-0.7$\%. As shown in Fig.~\ref{fig:fig_bar}, in $8$ of the $11$ models, the SA-score for T is higher than the DA-score, while the SA-score for V is lower than the DA-score in $8$ models. This suggests that in some models, discrete attribution based solely on the final generated options may underestimate the relative support of the model for the T-type answer.

The numerical differences in T-type attribution are mainly concentrated in models where the discrete TMB percentage is between $6$\% and $18$\%. For the two models with a TMB percentage below $3$\%, the differences between the two methods are no more than $0.3$\%. Specifically, the T-type soft attribution result for Qwen3.8-$27$b is $5.8$\%, while the discrete attribution result is $1.8$\%, a difference of $4.0$\%.

The above differences reflect the complementarity of the two methods: DA determines the information source based on the options finally generated by the model, while SA uses the probabilistic quality of the Top-K candidate tokens to characterize the relative support of different information sources at the candidate answer level. Therefore, SA can supplement the candidate distribution information that DA fails to reflect. However, it should be noted that the SA-score only reflects the normalized relative support ratio of the probability quality of the preset options that can be matched in the Top-K, and does not represent the probability distribution on the complete vocabulary.

\section{Conclusion}\label{sec5}

This paper presents a systematic investigation of information source attribution in VLMs under cross-modal information conflicts. We identify a critical distinction between TMB and LP, two fundamentally different sources of multimodal failures that can produce seemingly similar responses while reflecting distinct information dependencies. To distinguish these effects, we introduce a VTK-based attribution framework that explicitly separates V, T, and K. Building on this framework, we construct a controlled benchmark with carefully designed question–answer and image configurations, incorporating dual baseline verification to establish the prerequisites for reliable attribution. We further develop complementary discrete and soft attribution metrics to characterize model decisions through final response selection and the relative probability support captured by Top-K token log probabilities, respectively. Experiments across $11$ VLMs reveal a consistent dominance of LP over TMB: LP is the leading attribution category in every evaluated model and exceeds TMB by more than 10 percentage points in $10$ of them, highlighting the importance of distinguishing prior-driven responses from external text-induced failures when investigating multimodal hallucinations. Further analyses reveal distinct scaling patterns within the Qwen3-VL family, where visual attribution generally increases and TMB declines with model size, whereas LP follows a non-monotonic trajectory. Comparisons between Qwen2.5-VL and Qwen3-VL models further reveal differences in their reliance on visual, textual, and prior information across model generations. Finally, discrete and soft attribution exhibit strong rank consistency across models, while their quantitative discrepancies provide complementary perspectives on information-source attribution beyond final response labels alone. Collectively, this work establishes a unified framework and benchmark for systematically examining information dependencies in VLMs and offers a more fine-grained perspective on the origins of multimodal hallucinations.

% \section{Limitations}\label{sec6}

\textbf{Limitations.} Due to the complexity of the dataset's image construction, the current dataset contains only $300$ samples ($600$ AE measurements). The coverage of common sense domains and counterintuitive operation types is coarse-grained, and a systematic fine-grained labeling system has not yet been established. This limits the statistical power of domain-specific and operation-type-specific differential analysis and makes it difficult to support conclusions for any specific common sense category. Future work will combine the ViLP construction method with the expansion process described in Section 3.2 and 3.3 to continuously expand the data, prioritizing the addition of long-tail domain and composite counterintuitive samples.

% \begin{appendices}

% \section{Section title of first appendix}\label{secA1}

\section*{Acknowledgments}
This work was supported in part by the National Key R\&D Program of China under Grant 2025YFA1510900, by the National Natural Science Foundation of China under Grant 62673439, by the Key Technology Research and Development Program Project of Hangzhou (No. 2025SZD1A41), by the Baima Lake Laboratory Joint Fund of the Zhejiang Provincial Natural Science Foundation of China under Grant LBMHZ25F020002, and by the Xinmiao Project of the Zhejiang Provincial Applied Basic Research Program under Grant 2026XMGD024.

\section*{Data Availability}

The evaluation datasets used in this study were developed based on the publicly available ViLP dataset. During dataset adaptation, a subset of the original questions was modified, additional question fields were introduced, and identified mismatches were corrected. The resulting samples were further paired to construct the evaluation datasets. The resulting VTK-Bench dataset will be made publicly available upon publication of this article.



% \end{appendices}


\bibliography{sn-bibliography}% common bib file

\end{document}